# Online Monotone Optimization

Ian Gemp
College of Information and Computer Sciences
University of Massachusetts Amherst
Amherst, MA 01003
imgemp@cs.umass.edu

Coauthor Sridhar Mahadevan
College of Information and Computer Sciences
University of Massachusetts Amherst
Amherst, MA 01003
mahadeva@cs.umass.edu

August 28, 2016

**Abstract**

This paper presents a new framework for analyzing and designing no-regret algorithms for dynamic (possibly adversarial) systems. The proposed framework generalizes the popular online convex optimization framework and extends it to its natural limit allowing it to capture a notion of regret that is intuitive for more general problems such as those encountered in game theory and variational inequalities. The framework hinges on a special choice of a system-wide loss function we have developed. Using this framework, we prove that a simple update scheme provides a no-regret algorithm for monotone systems. While previous results in game theory prove individual agents can enjoy unilateral no-regret guarantees, our result proves monotonicity sufficient for guaranteeing no-regret when considering the adjustments of multiple agent strategies in parallel. Furthermore, to our knowledge, this is the first framework to provide a suitable notion of regret for variational inequalities. Most importantly, our proposed framework ensures monotonicity a sufficient condition for employing multiple online learners safely in parallel.

## 0.1 Introduction

Online optimization frameworks developed for online learning have enabled the design of simple, efficient, no-regret (i.e., sub-linear regret) algorithms for solitary agents in potentially adversarial environments. These frameworks elegantly capture theoretical properties of the online learning problem and provide practitioners with scalable solutions to a number of real-world problems (e.g., online advertising), however, none of these frameworks was expressly designed to approach the online learning problem for groups of agents or dynamical systems in general. While results from current online optimization frameworks naturally provide unilateral regret bounds for singular agents by treating all other agents as part of the environment, these bounds are not adequate for answering important questions such as, "How poor is my team of online learners with respect to the best team"? This question is especially important to distributed system design in possibly adversarial environments and arises in a variety of research fields such as game theory, multi-agent RL, variational inequalities, machine learning, and economics.

The famous Blackwell Approachability Theorem [5] provides conditions under which an algorithm exists ensuring the individual targets (e.g., rewards) of multiple agents approach a convex set $\mathcal{Z}$. Specifically, if after observing an adversary's play at each time step there exists a set of agent strategies that results in a set of targets $z \in \mathcal{Z}$, then there exists an algorithm producing targets whose average in the limit is at a distance zero from $\mathcal{Z}$. While this result appears to fill the gap described above, it has recently been shown that Blackwell Approachability is equivalent to online linear optimization (OLO) in a strong sense [1][1]. Furthermore, the associated approachability algorithm requires access to a halfspace oracle or potential function which assumes some level of cooperation among agents. In this work, we focus on developing a framework for designing distributed (or selfish) algorithms and so this is inadequate for our needs.

In developing a new framework we will need to restrict ourselves to domains satisfying specific properties that allow the design of simple, efficient, no-regret algorithms. Therefore, the study of actually constructing appropriate domains, known as mechanism design in game theory and economics, is intimately related to our work [4]. A mechanism is incentive compatible if individual agent rewards are maximized when acting according to true preferences. In other words, a mechanism is incentive compatible if the goal (e.g., system-wide honesty) is compatible with the incentivized behavior (e.g., profit maximization). In this work, the system-wide goal is low aggregate team loss and the incentivized behavior is selfish loss minimization.

This work makes two contributions. The first is an online learning framework for monotone maps, namely online monotone optimization, which generalizes the popular framework of online convex optimization and provides a foundation for algorithm design in the multi-agent setting, variational inequality setting, and others. The second is a simple, efficient, no-regret, distributed algorithm suitable for the newly developed framework. We begin with a brief background on convex analysis, monotone maps, variational inequalities, and online convex optimization (OCO). Section 2 presents the new online monotone optimization framework (OMO) along with a proof showing OMO is a strict superset of OCO. We also discuss in detail the meaning behind various components of the framework. Section 3 derives a simple, efficient, no-regret algorithm for the proposed framework and Section 4 demonstrates our contribution in solving a dynamic variational inequality problem. We finish with a discussion of results and future work.

[1] OLO $\subset$ OCO

# 1 Technical Background

First, we'll provide some useful tools and results from convex analysis, monotone operator theory, variational inequalities, and online convex optimization.

We denote the norm of a vector $x$ by $||x||$ and its dual norm by $||x||_* = \max\{\langle x', x\rangle : ||x'|| \leq 1\}$. A set is convex if $tx+(1-t)x' \in \mathcal{X}\ \forall x, x' \in \mathcal{X}, t \in [0,1]$. All sets we consider are convex.

The subdifferential of a function $f : \mathcal{X} \subset \mathbb{R}^n \rightarrow \mathbb{R}$ at $x$, denoted $\partial f(x)$, is the set of all subgradients at $x$: $\partial f(x) = \{z : \langle z, x'-x\rangle \leq f(x') - f(x)\ \forall x' \in \mathcal{X}\}$. $f$ is L-Lipschitz over $\mathcal{X}$ if $|f(x') - f(x)| \leq L||x'-x||$ or equivalently if $||z||_* \leq L\ \forall x, x' \in \mathcal{X}, \forall z \in \partial f(x)$. $f$ is convex if $\langle z - z', x - x'\rangle \geq 0\ \forall x, x' \in \mathcal{X}, z \in \partial f(x), z' \in \partial f(x')$[2].

A set valued map $F : \mathcal{X} \subset \mathbb{R}^n \rightarrow \{z \in \mathcal{Z} \subset \mathbb{R}^n\}$ is Lipschitz over $\mathcal{X}$ if $||F(x') - F(x)|| \leq L||x'-x||\ \forall x, x' \in \mathcal{X}$ or equivalently if $||z||_* \leq L\ \forall x, x' \in \mathcal{X}, \forall z \in F(x)$. $F$ is monotone if $\langle z - z', x - x'\rangle \geq 0\ \forall x, x' \in \mathcal{X}, z \in F(x), z' \in F(x')$ [3]. Alternatively, if $F$ is differentiable and the Jacobian of $F$ is positive semi-definite, $F$ is monotone [12, 17]. By our defintions, $F$ is guaranteed to be integrable over finite length paths through the domain $\mathcal{X}$. $F$ is conservative if the integral along any closed contour in $\mathcal{X}$ is zero: $\oint \langle F, dx\rangle = 0$[3]. Equivalently, $F$ is conservative if it is path-independent meaning the integral along a contour between any two points in $\mathcal{X}$ doesn't depend on the contour itself. Conversely, if $F$ is path-dependent or $\exists x \in \mathcal{X}$ such that $\oint \langle F, dx\rangle \neq 0$, then $F$ is not conservative.

Clearly, by definition, the subdifferential of a convex function is a monotone set valued map, but the converse is not necessarily true. If $F$ is monotone and conservative, then $F$ is the subdifferential of some convex function [16], however, not all monotone maps are conservative. We provide one such example in the next section. We emphasize this point to convince the reader that monotone maps are capable of representing problems not readily captured by convex functions. A more thorough discussion relating monotone maps to subdifferentials can be found in [16].

In Section 4 we demonstrate our framework on a variational inequality (VI) problem so we introduce VIs here. The variational inequality problem, VI$(F, \mathcal{X})$, is to find $x^*$ such that $\langle F(x^*), x - x^*\rangle \geq 0\ \forall x \in \mathcal{X}$. VI's are used to model equilibrium problems in a number of domains including mechanics, traffic networks, economics, and game theory. In our notation, $x^*$ constitutes an equilibrium point. It is known that the solution set, $\mathcal{X}^*$, to VI$(F, \mathcal{X})$ with monotone $F$ is a convex, compact set [6]. We refer the reader to [15] for a

[2]convexity is typically presented as $f(tx + (1-t)x') \leq tf(x) + (1-t)f(x')$, but the form we present better suits the transition to OMO.

[3]colloquially known as the fundamental theorem of calculus for line integrals.

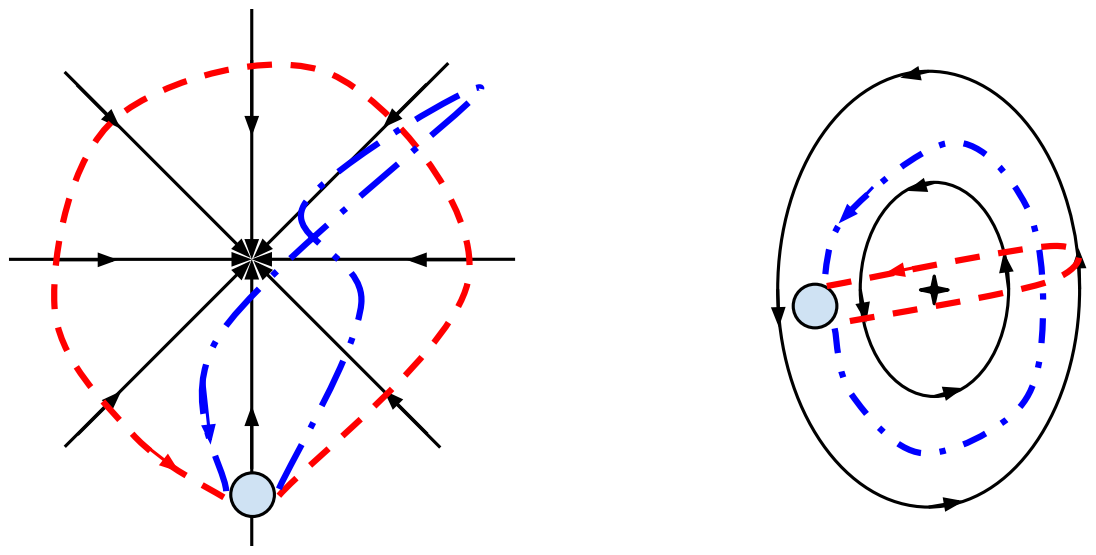

Figure 1: The sketch on the left represents the gradient map of a convex function (e.g., $x^2 + y^2$) while the sketch on the right is of a circular vector field. Both are monotone maps, however, the blue and red contour integrals in the left sketch are over a conservative map and so they both evaluate to zero (path-independent). In contrast, the blue and red contour integrals in the right sketch are over a non-conservative monotone map and possibly evaluate to different values (path-dependent).

detailed study of the relationship between convex functions, monotone operators, and variational inequalities and to [11, 12, 8] for an extensive study of VIs. Geometrically, monotonicity implies $\mathcal{X}^*$ is a global monotone attractor [12]. Figure 1 demonstrates a few of the aforementioned properties of monotonicity as it may be a new concept to some readers.

Online convex optimization (OCO) is a framework for studying the online learning problem when losses are convex with respect to the prediction domain which is also a convex set. The learning problem is defined in Algorithm 1. We refer the reader to the survey by Shalev-Shwartz [18] for a review of OCO. The standard no-regret algorithm for OCO is online gradient (mirror) descent; online gradient descent (OGD) is described in Algorithm 2. The regret of an online learning algorithm $\mathcal{A}$ predicting the sequence $x_t$ after $T$ steps is formally defined as $regret_{\mathcal{A}_T}(\mathcal{X}) = \sum_{t=1}^{T} f_t(x_t) - min_{u \in \mathcal{X}} \sum_{t=1}^{T} f_t(u)$. Assuming each $f_t$ is $L_t$-Lipschitz and $\mathcal{X} = \{x : ||x||_2 \leq B\}$, it has been shown that $regret_{OGD_T} \leq BL\sqrt{2T}$ where $L^2 \geq \frac{1}{T} \sum_{t=1}^{T} L_t^2$ [18].

# 2 New Framework

We are now ready to present our online monotone optimization framework (Algorithm 3). Here we denote $x : o_t \rightarrow x_t$ as $x$ taking the straight line path from $o_t$ to $x_t$ through $\mathcal{X}$ and $x : o_t \rightsquigarrow x_t$ as taking an arbitrary path (continuous, finite length). Comparing OMO to OCO, we see that the major difference is that we now receive a loss function implicitly defined by a monotone map whereas in OCO, we receive a convex loss function directly. We will now show that OMO is strictly more general than OCO.

**Theorem 1.** *$OCO(f_t, X)$ is equivalent to $OMO(\partial f_t, \mathcal{X})$ and $\exists F_t$ such that*

**Algorithm 1** Online Convex Optimization (OCO)

input: A convex set $\mathcal{X}$
**for** $t = 1, 2, \ldots$ **do**
predict a vector $x_t \in \mathcal{X}$
receive a convex loss function $f_t : \mathcal{X} \to \mathbb{R}$
suffer loss $f_t(x_t)$
**end for**

**Algorithm 2** Online Gradient Descent (OGD)

input: A scalar learning rate $\eta > 0$
$x_1 = 0$
**for** $t = 1, 2, \ldots$ **do**
$x_{t+1} = x_t - \eta z_t$ where $z_t \in \partial f_t(x_t)$
**end for**

**Algorithm 3** Online Monotone Optimization (OMO)

input: A convex set $\mathcal{X}$
**for** $t = 1, 2, \ldots$ **do**
predict a vector $x_t \in \mathcal{X}$
receive a monotone mapping $F_t : \mathcal{X} \to \mathcal{X}$ with reference vector, $o_t$, and reference scalar, $f_{o_t}$
suffer loss $f_t(x_t) = f_{o_t} + \int_{x:o_t \to x_t} \langle F_t, dx \rangle$
**end for**

*$OMO(F_t, \mathcal{X}) \notin OCO(f_t, \mathcal{X})$ $\forall f_t$ implying $OCO \subset OMO$ in the strict sense.*

*Proof.* First, we show that OMO simplifies to OCO when the monotone map $F_t$ is restricted to be conservative. We know from Section 1 that $F_t$ is the subdifferential of some convex function (to within a constant). Let $f_t$ be the convex function whose subdifferential is $F_t$ and whose value at $o_t$ is $f_{o_t}$ (i.e., $f_t(o_t) = f_{o_t}$). We also know that a conservative map is path-independent. The following steps illustrate the reduction to OCO by existence of the subdifferential, path-independence, and canceling integration constants. This is simply a straightforward application of the fundamental theorem of calculus for line integrals.

$$f_{o_t} + \int_{x:o_t \to x_t} \langle F_t, dx \rangle = f_{o_t} + \int_{x:o_t \to x_t} \langle \partial f_t, dx \rangle \quad (1)$$

$$= f_{o_t} + \int_{x:o_t \rightsquigarrow x_t} \langle \partial f_t, dx \rangle \quad (2)$$

$$= f_{o_t} + f_t(x_t) - f_t(o_t) \quad (3)$$

$$= f_t(x_t) \quad (4)$$

In other words, to execute OCO under the OMO framework, simply pass the subdifferential of the convex function as the monotone map and everything else remains the same.

Now that we know OMO contains OCO, we'll prove OMO is a strict superset of OCO by identifying an element of OMO that lies outside OCO. The counterexample we provide is a 2 player, 2 action game where the loss functions seen by both players at each time step $t$ are kept constant. Player 1 attempts to minimize the loss $\frac{1}{3}r_t^3 - \frac{1}{2}r_t^2 c_t + r_t c_t^2$ by adjusting $r_t \in [0,1]$, and player 2 attempts to minimize the loss $\frac{1}{3}c_t^3 - \frac{1}{2}c_t^2 r_t + c_t r_t^2$ by adjusting $c_t \in [0,1]$. The corresponding map is just the concatenation of their gradients, $F_t(x_t) = \binom{r_t^2 - r_t c_t + c_t^2}{r_t^2 - r_t c_t + c_t^2}$ with $x_t = \binom{r_t}{c_t}$ and $\mathcal{X} = [0,1]^2$. The Jacobian of $F_t$ is $\begin{pmatrix} 2r_t - c_t & 2c_t - r_t \\ 2r_t - c_t & 2c_t - r_t \end{pmatrix}$, which is positive semi-definite over $\mathcal{X}$, hence monotone. Let $o_t = (1,1)$ and $f_{o_t} = 0$.

We will evaluate $f_t(x)$ under the OMO framework and show it is non-convex for this counterexample proving this problem is outside the reach of OCO yet still within reach of OMO.

$$f_t(x) = f_{o_t} + \int_{x:o_t \to x_t} \langle F_t, dx \rangle \quad (5)$$

$$= \int_0^1 \langle F_t(o_t + \tau(x_t - o_t)), (x_t - o_t) d\tau \rangle \quad (6)$$

$$= \frac{1}{3}r_t^3 - \frac{1}{2}r_t^2 + r_t c_t - \frac{1}{2}c_t^2 + \frac{1}{3}c_t^3 - \frac{2}{3} \quad (7)$$

Hessian$(f_t(x)) = \begin{pmatrix} 2r_t - 1 & 1 \\ 1 & 2c_t - 1 \end{pmatrix}$ with $Det \leq 0$ over $\mathcal{X}$ which implies $f_t(x)$ is a saddle surface. □

From this viewpoint, OMO can be considered a specific version of online non-convex optimization. While there has been work on developing no-regret algorithms for specific non-convex formulations [19], our focus is on extending OCO to its limit. In other words, we argue monotonicity rather than convexity is the natural boundary for developing simple, efficient online optimization algorithms which will be apparent in our algorithm design.

## 2.1 Discussion of Losses and Regret in OMO

The OMO loss is implicitly defined by the monotone map $F_t$. In physics, this loss would be recognized as the work required to move a particle from $o_t$ to $x_t$ along a straight line path through field $F_t$. With respect to the multi-agent example given above, this loss represents the aggregate reward gained by both agents when linearly converting their strategies from $x_t$ to $o_t$. This is because the map $F_t$ is simply the concatenation of the gradients from each agent's expected reward function. In this scenario, we can rewrite our loss as $f_t(x_t) = \sum_{i=1}^N \int_{x:o_t \rightarrow x_t} \langle \partial V_t^{(i)}(x), dx^{(i)} \rangle$ where $i$ ranges over the agents $1, \ldots, N$ and $\partial V_t^{(i)}(x)$ is the subdifferential of agent $i$'s expected reward function with respect to its strategy $x_t^{(i)}$ evaluated at $x$. At this point, straight line paths may seem an arbitrary choice for the contour, so we explain this choice in detail.

One reason for defining our loss with straight line paths is that it upper bounds an arguably more intuitive loss we are interested in. First, we define the straight line path, $\vec{P}$, explicitly as $\vec{P}(o_t, x_t) = arg\min_{\tilde{P}(o_t,x_t)} \int_{\tilde{P}(o_t,x_t)} \alpha ||\tilde{P}||$, $\alpha \in (0, 1]$ where $\tilde{P}$ represents any arbitrary path and $||\tilde{P}||$ represents the length of the path. This is the classic example from variational calculus that a straight line is the shortest path between two points. Next, we define the least cost, myopic path as $P^*(o_t, x_t) = arg\min_{\tilde{P}(o_t,x_t)} \int_{\tilde{P}(o_t,x_t)} \alpha||\tilde{P}|| + (1 - \alpha)\langle F_t, d\tilde{P} \rangle$[4]. We say least cost, myopic because this formula returns the path that minimizes the work required to transition from $o_t$ to $x_t$ while penalizing long paths. In the context of profit maximization in economics, if we were to execute this transition in realtime, this trajectory would save us a large amount of money but still complete in a reasonable amount of time. It is clear by the definitions that

$$\int_{P^*(o_t,x_t)} \langle F_t, dP^* \rangle \leq \int_{\vec{P}(o_t,x_t)} \langle F_t, d\vec{P} \rangle, \tag{8}$$

and so any algorithms that minimize straight line paths also bound least cost, myopic paths.

Next, we revisit the standard definition for regret. While it's still possible to define regret in the same way as OCO, an alternative definition of regret arises in the OMO framework that is arguably more appealing. Both notions are equivalent in the OCO setting. Please consult Figure 2 for a visual during the following discussion. Standard regret compares the average losses of the online learning strategy to the best fixed strategy in retrospect ($\int_{o_t}^{x_t} - \int_{o_t}^{u_T}$). Another form of regret could be the loss felt when converting the best fixed strategy to the online learning strategy ($\int_{u_T}^{x_t}$). While these two notions are equivalent in OCO, they are possibly different in OMO due to path-dependence of general monotone maps. We will argue for our new notion of regret as it retains a crucial property. First we provide a proof sketch of a useful integral upper bound over monotone maps (see [16] for formal proof). Let $x_{i+1} - x_i = \frac{x_n - x_0}{n} \; \forall \; x_0, x_n$[5] and recall the definition of montonicity, $\langle F(x_{i+1}) - F(x_i), x_{i+1} - x_i \rangle \geq 0 \; \forall \; x_i, x_{i+1}$ which implies

$$\langle F(x_i), \frac{x_n - x_0}{n} \rangle \leq \langle F(x_{i+1}), \frac{x_n - x_0}{n} \rangle \; \forall \; x_i, x_{i+1} \tag{9}$$

$$\implies \langle F(x_i), \frac{x_n - x_0}{n} \rangle \leq \langle F(x_j), \frac{x_n - x_0}{n} \rangle \; \forall \; j \geq i \tag{10}$$

[4] $\alpha||\tilde{P}||$ is necessary because monotone maps admit cycles. If $F_t$ is Lipschitz, $\exists \alpha$ s.t. $||P^*||$ is finite.

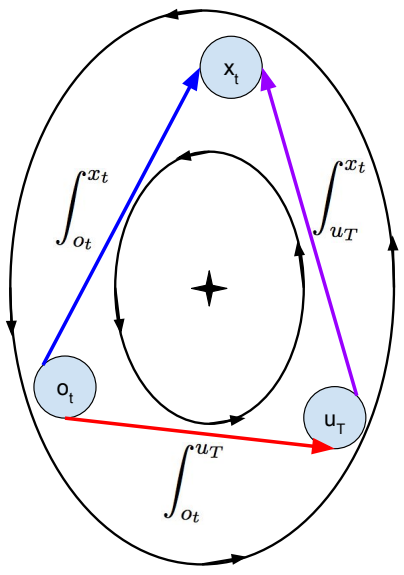

Figure 2: Illustrative comparison of alternative regret definition to standard defintion.

Also,

$$\langle F(x_0), x_n - x_0 \rangle = \langle F(x_0), \sum_{i=0}^{n-1} \frac{x_n - x_0}{n} \rangle \tag{11}$$

$$= \sum_{i=0}^{n-1} \langle F(x_0), \frac{x_n - x_0}{n} \rangle \tag{12}$$

$$\leq \sum_{i=0}^{n-1} \langle F(x_i), \frac{x_n - x_0}{n} \rangle \tag{13}$$

$$= \int_{x: x_0 \rightarrow x_n} \langle F, dx \rangle \text{ as } n \rightarrow \infty, \tag{14}$$

and vice versa for the reverse direction, which implies

$$\implies \langle F(x_0), x_n - x_0 \rangle \leq \int_{x: x_0 \rightarrow x_n} \langle F, dx \rangle \leq \langle F(x_n), x_n - x_0 \rangle. \quad \square \tag{15}$$

Using this bound, we have that standard regret, $\text{regret}_s = \int_{o_t}^{x_t} - \int_{o_t}^{u_T} \leq \langle F(x_t), x_t - o_t \rangle - \langle F(o_t), u_T - o_t \rangle$. Likewise, our new regret, $\text{regret}_n = \int_{u_T}^{x_t} \leq$

[5]Subscript here simply differentiates vectors. It does not denote a time step.

$\langle F(x_t), x_t - u_T\rangle = \langle F(x_t), x_t - o_t\rangle - \langle F(x_t), u_T - o_t\rangle$. Unfortunately, there's not much we can say regarding the latter terms, $\langle F(x_t), u_T - o_t\rangle$, and so it's not clear if one of these regrets upper bounds the other. However, we can bound the difference between the two regrets using Stoke's theorem and bounds on $F$ and its derivatives. The difference between the two regrets is equal to the magnitude of the path integral around the triangle in Figure 2. By Stoke's theorem, $|\oint_{\partial\Sigma}\langle F, dx\rangle| = |\int_\Sigma \nabla \times F \cdot \partial\Sigma| \leq \max_\Sigma |\nabla \times F| \times \int_\Sigma \partial\Sigma \leq 3\sqrt{\frac{1}{2}(\beta^2 + L\gamma)} \cdot ||u_T - o_t|| \cdot ||x_t - u_T||$ where $L$, $\beta$, and $\gamma$ are bounds on $F$, the Jacobian of $F$, and a matrix of 2nd derivates respectively and $\Sigma$ ($\partial\Sigma$) is the 2-dimensional area (perimeter) formed by the path (see Appendix). The last step bounds the norm of the curl as well as the area of the triangle. From this bound, we can see that the closer $x_t$ is to $u_T$ or the better $o_t$ is made to approximate $u_T$, the smaller the difference.

Our new notion of regret, however, has the nice property that it is independent of the reference vector $o_t$ which will aid our derivation of no-regret algorithms in the next section. Moreover, we argue our new notion of regret is satisfactory even when the standard regret exceeds it. If $\text{regret}_n \geq \text{regret}_s$ then any algorithm that minimizes $\text{regret}_n$ minimizes $\text{regret}_s$ as well, so let's assume, on the contrary, that $\text{regret}_n < \text{regret}_s$ for some choice of vectors $o_t, x_t, u_T$ and map $F_t$. There exists a path from $o_t$ to $x_t$ such that when measuring the loss of $x_t$, $f_t(x_t)$, the standard notion of regret is then equal to our new notion. The proof is trivial since the two segment path from $o_t$ to $x_t$ through $u_T$ exactly gives us this result. In other words, if the standard regret is greater than the regret given by our new definition, it is because the standard regret was evaluated along suboptimal paths and a tighter regret exists which we have explicitly defined. Note that none of these issues arise in the case where $F_t$ is a conservative, monotone map as integrals are then path-independent. This discussion is critical for only the non-conservative case.

In the context of economics, our new notion of regret upper bounds the amount of money a group of firms should be willing to pay to adjust their strategies in retrospect. We summarize our new definition here for convenience.

$$\mathit{regret}_{\mathcal{A}_{(t,T)}}(\mathcal{X}) = \int_{x:u_T \to x_t} \langle F_t, dx\rangle \leq \langle F_t(x_t), x_t\rangle - \langle F_t(x_t), u_T\rangle \tag{16}$$

$$\mathit{regret}_{\mathcal{A}_T}(\mathcal{X}) = \sum_{t=1}^{T} \mathit{regret}_{\mathcal{A}_{(t,T)}}(\mathcal{X}) \tag{17}$$

$$u_T = \mathit{arg}\min_{u\in\mathcal{X}} \sum_{t=1}^{T} \int_{x:o_t \to u} \langle F_t, dx\rangle \tag{18}$$

# 3 Derivation of No-Regret Algorithms

Due to the simplicity of the regret bounds discussed in the previous section and the work previously done in OCO, the derivation of no-regret algorithms for

**Algorithm 4** Online Monotone Descent (OMoD)

input: A scalar learning rate $\eta > 0$
$x_1 = 0$
**for** $t = 1, 2, \ldots$ **do**
  $x_{t+1} = x_t - \eta z_t$ where $z_t \in F_t(x_t)$
**end for**

**Algorithm 5** Online Monotone Mirror Descent (OMoMD)

input: A link function $g : \mathbb{R}^n \to \mathcal{X}$
$x_1 = g(0)$
**for** $t = 1, 2, \ldots$ **do**
  $\theta_{t+1} = \theta_t - \eta z_t$ where $z_t \in F_t(x_t)$
  $x_t = g(\theta_{t+1})$
**end for**

OMO is trivial. We have shown that immediate regret for general monotone maps can be bounded above by considering a constant approximation of the map (see 16). Note that a constant map, $F_t$, is always the subdifferential of some linear function, $f_t = \langle F_t(x_t), x \rangle$. This implies that the regret for general monotone maps is bounded above by considering the online linear optimization problem with $f_t$. This reduction mirrors that of OCO, where convex losses are bounded above by their linear approximations. The implication is that the online gradient decent and even online mirror descent algorithms can be adapted from OCO with almost no effort to minimize regret in OMO while enjoying exactly the same $o(T)$ regret bounds (see end of Section 1 and supplementary material). This is somewhat surprising as we showed in Section 2 (see 33) that OMO sometimes involves minimizing non-convex functions. The no-regret algorithms for OMO are given in Algorithms 4 and 5.

As you can see, the only difference between the new algorithms and those designed for OCO are that we now allow monotone maps that are not necessarily subdifferentials of any function. This is further evidence that monotonicity rather than convexity is a more maximal condition for the development of simple, efficient, no-regret algorithms.

## 3.1 Existence of No-regret Algorithms for OMO

As stated above, OMO generally involves minimizing regret for non-convex functions. Even though this is a surprising result in and of itself, the counterexample we provided actually hides some of the real difficulty of the monotone optimization problem. In that example, we were able to calculate the definite integral in closed form, which is an artifact of the example's simplicity. In general, computing the best fixed strategy in retrospect (see 18) is by itself a difficult optimization problem because of the integral. Evaluating the loss as well as the gradient with respect to a strategy both involve integration. If the integral cannot be computed in closed form, then it must be integrated numerically which can be expensive depending on the size of the problem. Furthermore, the algorithms we present here only require the evaluation of $F_t$ at a single point $x_t$. Integration, numerical or otherwise, implies that the algorithm needs access to the map itself which is not always available. In the case where only $F_t(x_t)$ is available, our algorithms will still perform with sublinear regret.

## 4 Experiment

We demonstrate the OMO framework and algorithms on a game-theoretic model of a cloud-based machine learning network (MLN), which has players (i.e., firms) compete to maximize profits by adjusting the quantity (e.g., # of samples $\times$ # of features), quality (e.g., Latency), and price (e.g., \$) of data delivered (see Figure 3). Providers of machine learning data, or service providers, (e.g., Twitter, Wikipedia) play a *Cournot-Nash* game controlling the quantity of data provided while network providers (e.g., Verizon, AT&T) play a *Bertrand game* controlling the delivery price as well as service quality. Consumers (e.g., tech firms, industrial research labs, universities) influence the network through demand functions dictating the prices they are willing to pay for specific quantities and qualities of services rendered. See supplementary material for a more thorough description of the MLN plus a second demonstration on an environmentally conscious supply chain.

The firms in the network continuously adjust their respective service offerings, optimizing their utilities, until any unilateral adjustment attempted by one firm is inherently detrimental to that firm's utility function. We assume the governing equilibrium is Cournot-Bertrand-Nash and firm utility functions are all concave and fully differentiable. This establishes the equivalence between the equilibrium state we are searching for and the variational inequality to be solved where $F_t$ returns a vector consisting of the negative gradients of the utility functions for each firm. Since $F_t$ is essentially a concatenation of gradients arising from multiple independent, conflicting objective functions, it does not correspond to the gradient of any single objective function. This prevents us from phrasing this equilibrium problem readily as an optimization problem.

To cast this VI as an online learning problem, we allow the parameters of the network to change. This actually creates a more realistic model as a number of external factors can cause the network to change such as weather, complex network congestion effects, cyber attacks, etc. The goal then is to predict the equilibrium point of each new VI in the face of these possibly adversarial forces. Specifically, our experiment considers ten different five-firm networks with monotone $F_t$. At each time step, the adversary receives the algorithm's prediction for the equilibrium point and returns the VI whose equilibrium is farthest from the predicted one. The reference vectors in this case are the solutions to the VIs, $x_t^*$. The reference constants are all assumed to be zero without loss of generality. For obvious reasons, we call this problem an Online Monotone Equilibration (OME) problem.

As discussed in subsection 3.1, computing the optimal strategy $u_T$ is intractable in general. Assuming networks are sampled uniformly in the limit, we expect $u_T$ to be the equilibrium of the network created from averaging the ten together, and so we treat that equilibrium as $u_T$ throughout the experiment. We measure three losses throughout the learning process. The first, which is an approximation of our new regret ($\text{regret}_n$), is the straight line integral from the current strategy to $\bar{x}^*$. The second, which is an approximation of the standard immediate regret ($\text{regret}_s$), is the difference between straight line integral from the reference vector $o_t$ to the current strategy minus the straight line integral from $o_t$ to $\bar{x}^*$. The third is the loss that the system would accrue if the current network was frozen and the system was allowed to reach the equilibrium $x_t^*$ ($\text{loss}_\infty$). Figure 4 plots the average of these measures with respect to the time step. Clearly $\text{regret}_n$ is decaying towards zero in support of our derived sublinear bounds.

# 5 Conclusion & Future Work

We proposed a new framework for online learning, namely online monotone optimization, which enables the study of regret for monotone maps. This framework generalizes the popular online convex optimization framework in a way that allows it to model regret for multiple agents in parallel while still retaining the simplicity of standard no-regret algorithms from previous work. We support the efficacy of our new framework with empirical results from a economic game model.

While we did not discuss it, this work is closely related to that of projected dynamical systems research. In non-adversarial settings, algorithms developed for deterministic or stochastic differential equations will better serve the problem. Our framework is designed to analyze the worst case scenario.

It is interesting to note that OMO has a strong relationship to online linear optimization just like Blackwell's Approachability framework and OCO. We are curious to learn if there is a property more general than monotonicity that is similarly capable of reducing to a bilinear form.

In future work, we hope to develop the framework further to handle dynamic regret as well as pseudo-montone maps and certain non-monotone maps. We also hope to develop efficient algorithms for approximating the optimal strategy $u_T$ in retrospect. Finally, recent work in adversarial networks [10, 7, 2, 9] inspires us to design monotone adversarial networks that could be trained online.

# 6 Appendix

## 6.1 OMoD & OMoMD Regret Bounds

We repeat the bounds adopted from [18] for convenience.

**Theorem 2.** *Let $R$ be a $(1/\eta)$-strongly-convex function over $\mathcal{X}$ with respect to a norm $||\cdot||$. Assume that OMoMD is run on the sequence of monotone maps, $F_t$, with the link function*

$$g(\theta) = arg\max_{x\in\mathcal{X}}(\langle x, \theta\rangle - R(x)). \tag{19}$$

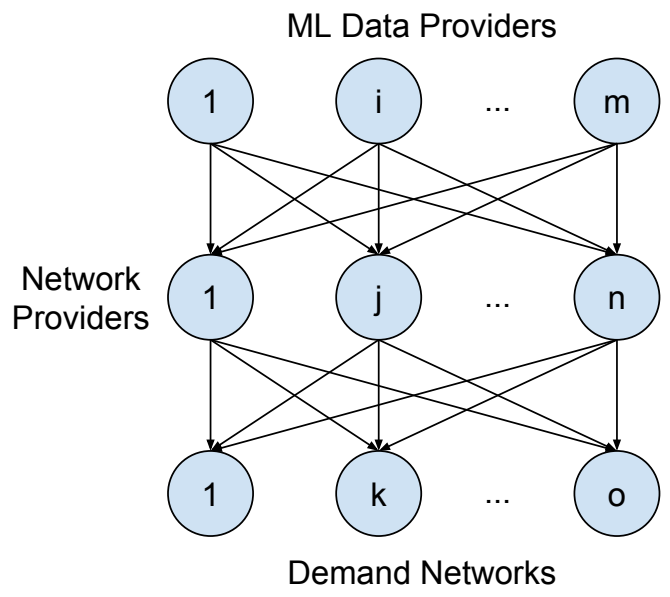

Figure 3: A next-generation economic model of a cloud based machine learning network (MLN) adapted from the service oriented internet model proposed in [13].

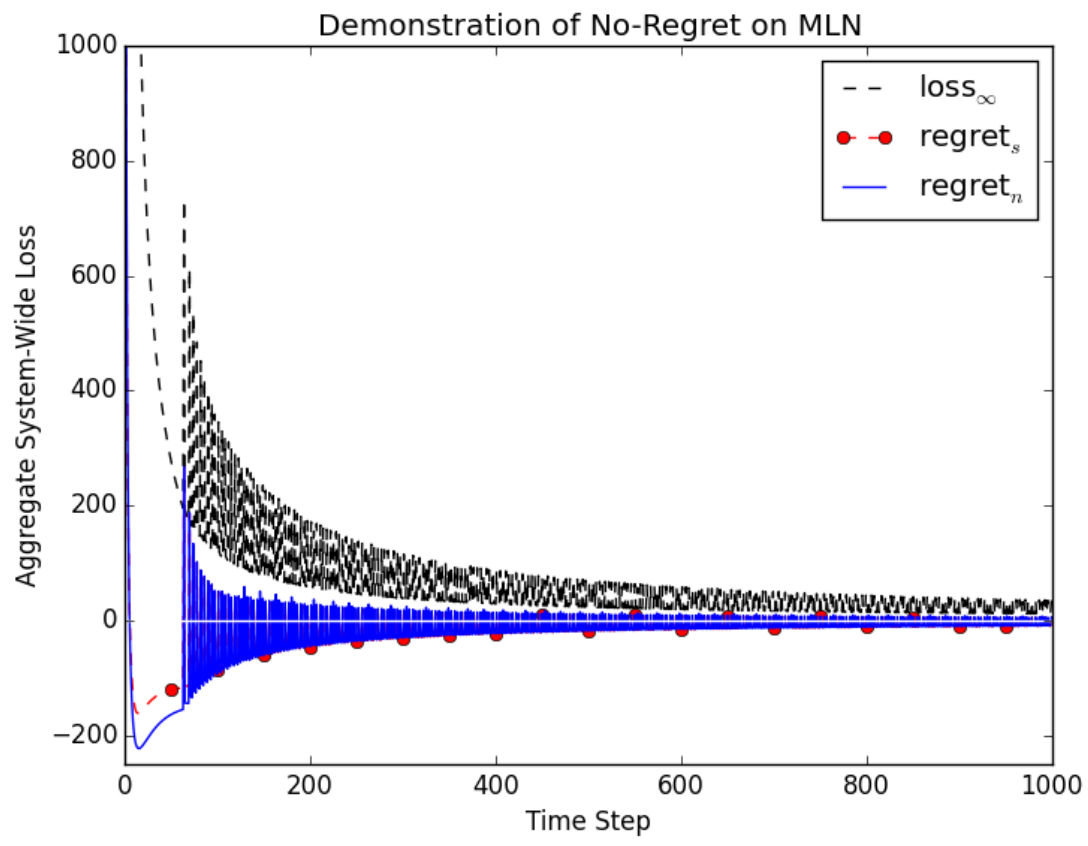

Figure 4: Demonstration of OMoMD on described machine learning network.

*Then, for all $u \in \mathcal{X}$,*

$$regret_{OMoMD_T}(\mathcal{X}) \leq R(u_T) - \min_{v \in \mathcal{X}} R(v) + \eta \sum_{t=1}^{T} ||z_t||_*^2 \tag{20}$$

*Furthermore, if $F_t$ is $L_t$-Lipschitz with respect to $||\cdot||$, then we can further upper bound $||z_t||_* \leq L_t$.*

*Proof.* As we have shown previously,

$$regret_{\mathcal{A}_{(t,T)}}(\mathcal{X}) = \int_{x:u_T \to x_t} \langle F_t, dx \rangle \leq \langle F_t(x_t), x_t \rangle - \langle F_t(x_t), u_T \rangle \tag{21}$$

and the OMoMD algorithm is equivalent to running Follow the Regularized Leader (FoReL) on the sequence of linear functions with the regularization $R(x)$. The theorem now follows directly from Theorem 2.11 and Lemma 2.6 in [18]. □

OMoD is equivalent to OMoMD with $R(x) = \frac{1}{2}||x||_2^2$ and so the proof for OMoMD extends to OMoD as well.

## 6.2 Machine Learning Network Motivation

The example in the paper demonstrates our proposed no-regret algorithm on a cloud-based machine learning network. Our network is motivated by expectations of the next era of machine learning. Data is often the difference between a high performing model and a mediocre one; for some data hungry models (e.g., deep learning), *Big Data* launches them to state-of-the-art results. We expect *Big Data* to drive a mature digital supply chain capable of supporting an economy where producers provide data for consumers (i.e., machine learning models) to consume. Unlike the present, this commodity will not be transferred into local storage for consumption on personal machines; rather, it will be transmitted in batches, immediately consumed for training, and discarded to allow room for the next batch. Our model of a cloud-based machine learning network (MLN) is trivially adapted from the service oriented internet (SOI) model proposed in [13]. In the original SOI model, service providers (e.g., Netflix, Amazon) stream content (e.g., movies, music). In our MLN model, service providers (e.g., Twitter, Wikipedia) stream machine learning data. Service providers control the quantity of data (i.e., # of samples × # of features) flowing through the market. Network providers charge service providers a fee for transmitting their data to consumers. The price different consumer markets are willing to pay service providers to stream data over a network of a certain quality is given by demand functions, price(quantity,quality). Given these relationships, service providers and network providers attempt to maximize their profits by varying their respective controls (quantity,quality) over the network. These relationships are parameterized so that we can instantiate ten five-firm networks by drawing parameters from uniform distributions over predefined ranges (code on github.com/*****).

### 6.3 Supply Chain Experiment

Our second example focuses on an emissions-conscious competitive supply chain network [14]. In this network model, $I$ firms manufacture products which are then either transported directly to retailers (demand markets) or to storage facilities for later distribution. The products in this economy are substitutable and distinguishable only by brand (eg. oil). In addition, we assume knowledge of the demand functions stating the prices markets are willing to pay for quantities of each product. In Figure 5, the nodes from top tier to bottom tier represent the firms ($i$), manufacturing plants ($M^i_m$), storage warehouses ($D^i_{d,1}\&D^i_{d,2}$), and demand markets ($R_r$). Each link in the network represents a process acting on the product between the origin and destination nodes. From top tier to bottom tier, the links represent manufacturing, transportation, storage, and distribution. Note that each $D^i_{d,1}$ and $D^i_{d,2}$ pair actually represents the same distribution center. This is because storage is a process that starts and ends in the same warehouse, hence the duplication of the nodes.

Each firm must decide how to optimally deliver its product to consumers given the allowable paths from its firm to the multiple demand markets. They do this by controlling their product flows (eg. barrels of oil per day) and frequencies of operation (eg. shipments per day) along paths in the network subject to capacity constraints (eg. barrels per shipment). For example, firm 1 may decide on two paths to optimize its supply chain: each day, two 150-barrel shipments are produced at well 1 and transported using mode 4 (barge) directly to retail market 1 and six 20-barrel shipments are produced at well 1 as well but are then transported using mode 3 (truck) to warehouse 2 for storage until they are finally distributed to retail market 11.

The firms in the network continuously adjust their product flows and operation frequencies, optimizing their utilities, until any unilateral adjustment attempted by one firm is inherently detrimental to that firm's utility function. Rationally competing on the basis of product output is known as Cournot competition and the stalemate described is known as a Nash equilibrium hence this state is known as a Cournot-Nash equilibrium. Given each firm's utility function and capacity constraints, we aim to find the corresponding steady-state product flows and frequencies of operation.

We assume the governing equilibrium is Cournot-Nash and the utility functions are all concave and fully differentiable. This establishes the equivalence between the equilibrium state we are searching for and the variational inequality to be solved where the $F_t$ mapping is a vector consisting of the negative gradients of the augmented Lagrangian utility functions for each firm. Since $F_t$ is essentially a concatenation of gradients arising from multiple independent, conflicting objective functions, it does not correspond to the gradient of any single objective function. This prevents us from phrasing this problem using a standard optimization formulation.

The experimental setup is the same as that for the MLN (ten three-firm networks) with code available on github.com/*****. Figure 6 plots the the same measures as in the MLN demonstration. Clearly $\text{regret}_n$ is decaying towards

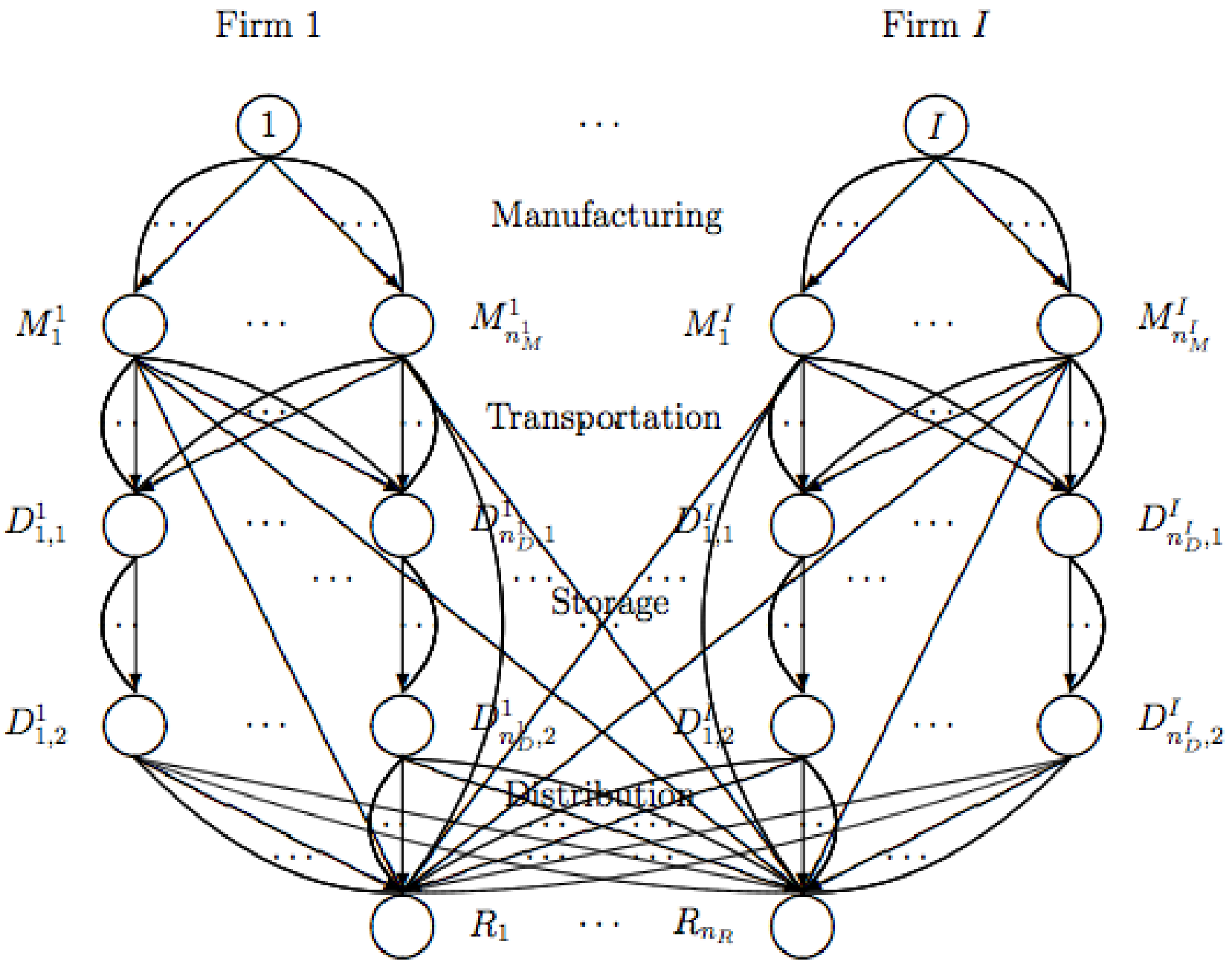

Figure 5: A "green" economic model of the supply chain proposed in [14]. Firms are modeled as playing a Cournot-Nash game, competing on the basis of product flow and frequency of operation. Demand markets consisting of individuals or groups of users choose between the various products offered by the firms.

zero in support of our derived sublinear bounds.

## 6.4 Counterexample Proof

### 6.4.1 $F_t$ is monotone over $\mathcal{X} = [0,1]^2$

$$F_t(x_t) = \begin{pmatrix} r_t^2 - r_t c_t + c_t^2 \\ r_t^2 - r_t c_t + c_t^2 \end{pmatrix} \tag{22}$$

$$Jacobian(F_t) = \begin{pmatrix} 2r_t - c_t & 2c_t - r_t \\ 2r_t - c_t & 2c_t - r_t \end{pmatrix} \tag{23}$$

$$= \begin{pmatrix} a & b \\ a & b \end{pmatrix} \text{ where } a = 2r_t - c_t \text{ and } b = 2c_t - r_t \tag{24}$$

$$\text{with eigenvalues } \lambda_{1,2} = 0, a + b \tag{25}$$

$$\lambda_{1,2} = 0, r_t + c_t \geq 0 \tag{26}$$

$$\implies Jacobian(F_t) \succeq 0 \implies F_t \text{ is monotone } \surd \tag{27}$$

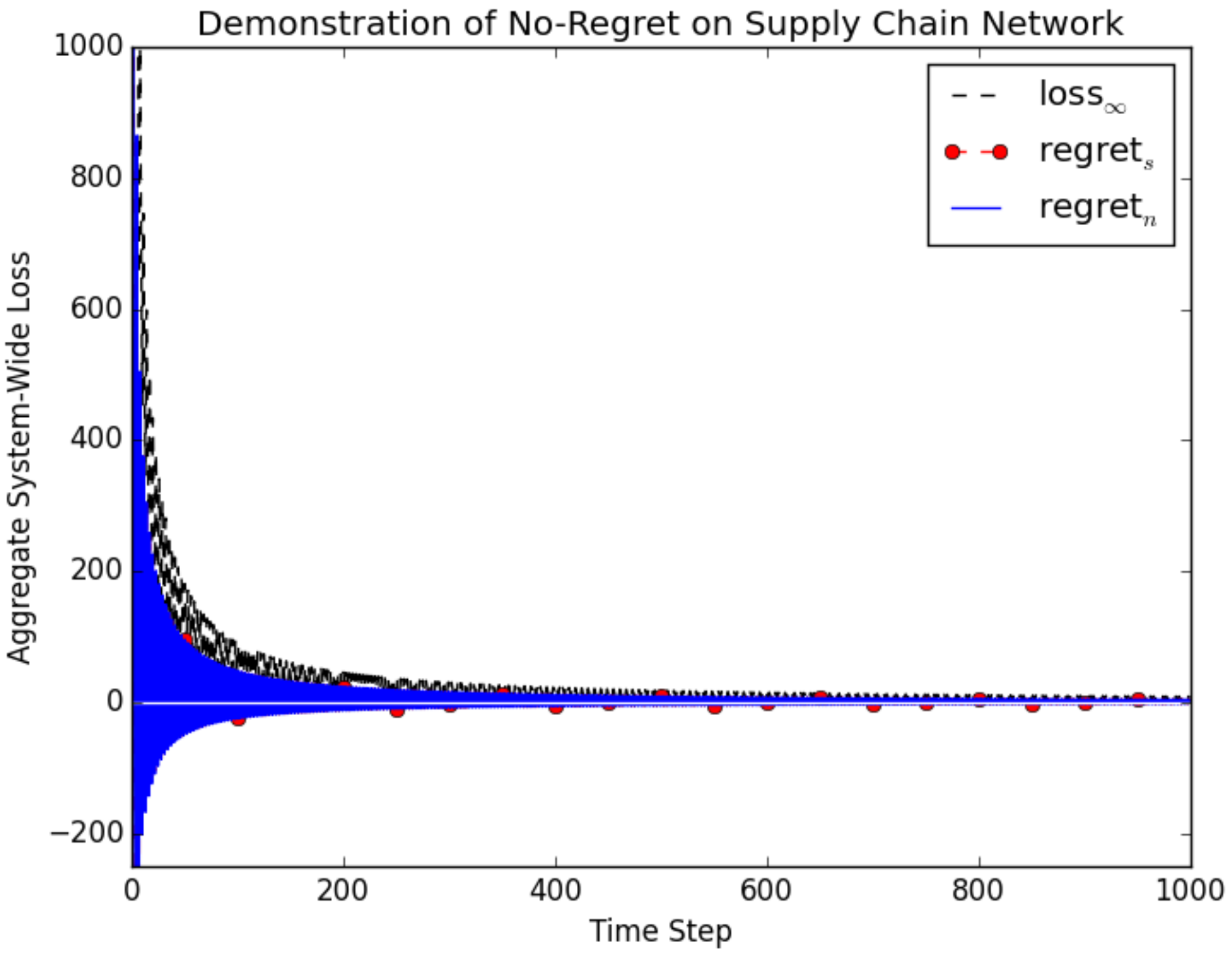

Figure 6: Demonstration of OMoMD on described machine learning network.

#### 6.4.2 $f_t$ is non-convex over $\mathcal{X} = [0,1]^2$

$$f_t(x) = f_{o_t} + \int_{x:o_t \to x_t} \langle F_t, dx \rangle \tag{28}$$

$$= \int_0^1 \langle F_t(o_t + \tau(x_t - o_t)), (x_t - o_t) d\tau \rangle \tag{29}$$

$$= \frac{1}{3}r_t^3 - \frac{1}{2}r_t^2 + r_t c_t - \frac{1}{2}c_t^2 + \frac{1}{3}c_t^3 - \frac{2}{3} \quad \text{see Mathematica code} \tag{30}$$

$$Hessian(f_t) = \begin{pmatrix} 2r_t - 1 & 1 \\ 1 & 2c_t - 1 \end{pmatrix} \tag{31}$$

$$= \begin{pmatrix} a & 1 \\ 1 & b \end{pmatrix} \text{ where } a = 2r_t - 1 \in [-1,1] \text{ and } b = 2c_t - 1 \in [-1,1] \tag{32}$$

$$Det(Hessian) = ab - 1 \leq 0 \implies f_t \text{ is a saddle surface} \tag{33}$$

#### 6.4.3 OMO $\equiv$ OCO for Positive Semi-Definite Affine Maps

This concerns such problems as linear complementarity problems (LCPs).

$$F_t(x_t) = Ax_t + b, \text{ where } x_t, b, o_t \in \mathbb{R}^n, \ A \succeq 0 \in \mathbb{R}^{n \times n} \tag{34}$$

$$f_t(x_t) = f_{o_t} + \int_{x:o_t \to x_t} \langle F_t, dx \rangle \tag{35}$$

$$= \int_0^1 \langle F_t(o_t + \tau(x_t - o_t)), (x_t - o_t) d\tau \rangle \tag{36}$$

$$= \int_0^1 \langle A(o_t + \tau(x_t - o_t)) + b, (x_t - o_t) d\tau \rangle \tag{37}$$

$$= \int_0^1 \langle Ao_t + \tau A(x_t - o_t) + b, (x_t - o_t) d\tau \rangle \tag{38}$$

$$= \int_0^1 \langle Ao_t + b, (x_t - o_t) d\tau \rangle + \tau \langle A(x_t - o_t), (x_t - o_t) d\tau \rangle \tag{39}$$

$$= \langle Ao_t + b, (x_t - o_t) \rangle + \frac{1}{2} \langle A(x_t - o_t)), (x_t - o_t) \rangle \tag{40}$$

$$= o_t^T A^T x_t - o_t^T A^T o_t + b^T (x_t - o_t) + \frac{1}{2}(x_t - o_t)^T A^T (x_t - o_t) \tag{41}$$

$$= o_t^T A^T x_t - o_t^T A^T o_t + b^T (x_t - o_t) + \dots \tag{42}$$

$$\frac{1}{2}[x_t^T A^T x_t - o_t^T A^T x_t - x_t^T A^T o_t + o_t^T A^T o_t] \tag{43}$$

$$= \frac{1}{2}[x_t^T A^T x_t + x_t^T (A - A^T) o_t - o_t^T A^T o_t] + b^T (x_t - o_t) \tag{44}$$

$$Hessian(f_t) = \frac{1}{2}[A + A^T] \succeq 0 \implies f_t \text{ is convex} \tag{45}$$

### 6.4.4 Curl Bound

$$J(F) = \begin{bmatrix} \frac{\partial F_1}{\partial x_1} & \frac{\partial F_1}{\partial x_2} & \frac{\partial F_1}{\partial x_3} \\ \frac{\partial F_2}{\partial x_1} & \frac{\partial F_2}{\partial x_2} & \frac{\partial F_2}{\partial x_3} \\ \frac{\partial F_3}{\partial x_1} & \frac{\partial F_2}{\partial x_2} & \frac{\partial F_2}{\partial x_3} \end{bmatrix} \tag{46}$$

$$J^2(F) = \begin{bmatrix} \frac{\partial^2 F_1}{\partial x_1^2} & \frac{\partial^2 F_1}{\partial x_2^2} & \frac{\partial^2 F_1}{\partial x_3^2} \\ \frac{\partial^2 F_2}{\partial x_1^2} & \frac{\partial^2 F_2}{\partial x_2^2} & \frac{\partial^2 F_2}{\partial x_3^2} \\ \frac{\partial^2 F_3}{\partial x_1^2} & \frac{\partial^2 F_2}{\partial x_2^2} & \frac{\partial^2 F_2}{\partial x_3^2} \end{bmatrix} \tag{47}$$

$$\rho(A) = \text{ largest singular value of } A \text{ which is square root of largest eigenvalue of } A^T A \tag{48}$$

$$F^R = R \cdot F \tag{49}$$

$$||RF||_2 \leq ||R||_2 ||F||_2 = ||F||_2 \text{ all matrix norms are submultiplicative and spectral norm of rotation matrix is 1} \tag{50}$$

$$F' = \text{ 3rd principal submatrix of } F^R \tag{51}$$

F' is a 3rd principal submatrix of rotation of F which means orthogonal projection after rotation
general principal submatrix spectral bound (with rotation)

Lemma 1

$$||A'||^2_{max} \leq ||RA||^2_{\max} \qquad \text{because principal submatrix just removing entries} \tag{52}$$

$$\leq ||RA||_2^2 \qquad ||A||_{\max} \leq ||A||_2 \tag{53}$$

$$\leq ||R||_2^2 ||A||_2^2 \qquad \text{lp induced norms submultiplicative} \tag{54}$$

$$\leq ||A||_2^2 \qquad \text{rotation matrix has unit spectral bound} \tag{55}$$

Lemma 2

$$||A'||_F^2 \leq 9||A'||^2_{max} \qquad \sum_{ij} |A_{ij}|^2 \leq n^2 \max(|A_{ij}|)^2 = n^2 ||A||^2_{\max} \tag{56}$$

$$\leq 9||A||_2 \qquad \text{by Lemma 1} \tag{57}$$

Lemma 3

$$||A'||_{2,1} = \sum_j ||a'_j||_2 \leq \sum_j ||a'_j||_1 \qquad ||a'||_2 \leq ||a'||_1 \tag{58}$$

$$\sum_j ||a'_j||_1 = \sum_{ij} |a'_{ij}| \qquad \text{by definition} \tag{59}$$

$$\sum_{ij} |a'_{ij}| \leq 9 \max_{ij} |a'_{ij}| = 9||A'||_{\max} \qquad \text{by inspection} \tag{60}$$

$$\leq 9||A||_2^2 = 9\lambda_{max}(A^T A) \qquad \text{by Lemma 1} \tag{61}$$

$$||\nabla \times F'(x)||_2^2 \leq ||\nabla||_2^2 ||F'(x)||_2^2 \qquad ||a \times b|| = ||a|| \cdot ||b|| \sin\theta \tag{62}$$

$$= \Delta ||F'(x)||_2^2 \qquad ||\nabla||_2^2 = \nabla \cdot \nabla = \Delta \tag{63}$$

$$= \Delta \sum_{i=1}^{3} F_i'(x)^2 \qquad ||F'(x)||_2^2 = F'(x) \cdot F'(x) \tag{64}$$

$$= \sum_{i=1}^{3} \Delta F_i'(x)^2 \qquad \text{linearity of } \nabla^2 \tag{65}$$

$$= 2 \sum_{i=1}^{3} \sum_{j=1}^{3} \frac{\partial F_i'(x)}{\partial x_j}^2 + F_i'(x) \frac{\partial^2 F_i'(x)}{\partial x_j^2} \qquad \text{evaluating } \Delta \tag{66}$$

$$= 2 \Big[ ||J(F'(x))||_F^2 + \sum_{j=1}^{3} \sum_{i=1}^{3} F_i'(x) \frac{\partial^2 F_i'(x)}{\partial x_j^2} \Big] \qquad \text{by definition} \tag{67}$$

$$= 2 \Big[ ||J(F'(x))||_F^2 + \sum_{j=1}^{3} F'(x) \cdot \frac{\partial^2 F'(x)}{\partial x_j^2} \Big] \qquad \text{write as dot product} \tag{68}$$

$$\leq 2 \Big[ ||J(F'(x))||_F^2 + \sum_{j=1}^{3} ||F'(x)||_2 \cdot || \frac{\partial^2 F'(x)}{\partial x_j^2} ||_2 \Big] \qquad a \cdot b = ||a|| \cdot ||b|| \cos\theta \tag{69}$$

$$= 2 \Big[ ||J(F'(x))||_F^2 + ||F'(x)||_2 \sum_{j=1}^{3} || \frac{\partial^2 F'(x)}{\partial x_j^2} ||_2 \Big] \qquad \text{commutivity} \tag{70}$$

$$= 2 \Big[ ||J(F'(x))||_F^2 + ||F'(x)||_2 ||J^2(F'(x))||_{2,1} \Big] \qquad \text{by definition} \tag{71}$$

$$\leq 2 \Big[ ||J(F'(x))||_F^2 + ||F(x)||_2 ||J^2(F'(x))||_{2,1} \Big] \qquad \begin{array}{c}\text{norm invariant to rotation} \\ \text{vector projection is coercive}\end{array} \tag{72}$$

$$\leq 2 \Big[ 9||J(F(x))||_2^2 + ||F(x)||_2 ||J^2(F'(x))||_{2,1} \Big] \qquad \text{by Lemma 2} \tag{73}$$

$$\leq 2 \Big[ 9\beta^2 + L||J^2(F'(x))||_{2,1} \Big] \qquad \text{by definitions} \tag{74}$$

$$\leq 2 \Big[ 9\beta^2 + 9L||J^2(F(x))||_2 \Big] \qquad \text{by Lemma 3} \tag{75}$$

$$\leq 18(\beta^2 + L\gamma) \qquad \text{by definitions} \tag{76}$$

$$||\nabla \times F'(x)||_2 \leq \sqrt{18(\beta^2 + L\gamma)} \qquad \text{take square root} \tag{77}$$